\documentclass[11pt]{article}
\usepackage{coling2016}
\usepackage[english]{babel}
\usepackage{times}
\usepackage{url}
\usepackage{latexsym}
\usepackage{comment}
\usepackage{csquotes}

\usepackage{epsfig}
\usepackage[colorinlistoftodos]{todonotes}
\usepackage{graphicx}
\usepackage{amsmath}
\usepackage{float}
\usepackage{algorithm,algpseudocode,caption}
\usepackage{subcaption}
\usepackage{paralist}
\usepackage{xspace}
\usepackage{url}
\usepackage[T1]{fontenc}
\usepackage{csquotes}
\usepackage{caption}

\usepackage{gb4e} 
\noautomath
\usepackage{caption} 
\usepackage{longtable} 

\title{Semantic Relation Classification: Task Formalisation and Refinement}

\author{Vivian S. Silva\textsuperscript{1}, Manuela H\"urliman\textsuperscript{2}, Brian Davis\textsuperscript{2}, \\ \textbf{Siegfried Handschuh\textsuperscript{1} and Andr\'{e} Freitas\textsuperscript{1}}\\
 \textsuperscript{1}Department of Computer Science and Mathematics, University of Passau, Passau, Germany \\
\textsuperscript{2} Insight Centre for Data Analytics, National University of Ireland, Galway, Ireland\\
 {\tt {\small \{vivian.santossilva, siegfried.handschuh, andre.freitas\}@uni-passau.de}} \\
 {\tt {\small\{manuela.huerlimann, brian.davis\}@insight-centre.org}}}

\date{}

\begin{document}
\maketitle

\begin{abstract}
The identification of semantic relations between terms within texts is a fundamental task in Natural Language Processing which can support applications requiring a lightweight semantic interpretation model. Currently, semantic relation classification concentrates on relations which are evaluated over open-domain data. This work provides a critique on the set of abstract relations used for semantic relation classification with regard to their ability to express relationships between terms which are found in a domain-specific corpora. Based on this analysis, this work proposes an alternative semantic relation model based on reusing and extending the set of abstract relations present in the DOLCE ontology. The resulting set of relations is well grounded, allows to capture a wide range of relations and could thus be used as a foundation for automatic classification of semantic relations.

\end{abstract}


\section{Introduction}\label{sec:intro}\blfootnote{This work is licenced under a Creative Commons Attribution 4.0 International Licence. Licence	details: http://creativecommons.org/licenses/by/4.0/}

The identification of abstract semantic relations between terms in text has emerged as a Natural Language Processing technique which is useful in a variety of tasks that depend on the extraction of key semantic relations from text.
 In essence, the task of semantic relation classification (SRC) consists in identifying common abstract relations such as causal, hypernymic and meronymic as relationships between terms in the text.

This definition puts semantic relation classification in the context of ontology extraction from text, where the emphasis is on the process of extracting more general and abstract relations, in contrast to more domain-specific relations.

However, despite the obvious intuition around the utility of the task, the justification on the scoping of the semantic relations set and their expressive coverage has not been fully grounded with regard to an ontological framework.
In contrast to this situation, the set of relations expressed within foundational ontologies are more formally axiomatised and built under conceptually well grounded methodologies.

Complementarily, the semantic relation classification task provides a corpus-based analysis on the incidence of these semantic relations on discourse, providing the fine-grained semantic context in which these abstractions are instantiated.
 However, when projecting these semantic relations back to the corpora-level, it can be observed that the majority of the words within a text does not have a direct semantic relationship connecting them.

Recent semantic interpretation tasks targeting word prediction over broader discourse contexts \cite{paperno2016} may require the detection of broader and complex semantic relations. Addressing these interpretation tasks may strongly benefit from relating terms expressed into the sentence using compositions of semantic relations.

This work aims at improving the description and the formalisation of the semantic relation classification task by grounding it with a foundational ontology and by introducing the concept of composite semantic relations, in which the relations between terms within a text can be expressed using the composition of one or more relations.

This work focuses on the following contributions:

\begin{itemize}
\item Examination of the completeness of the set of semantic relations used for the evaluation of semantic relation classification (SRC) tasks in the context of a domain-specific corpus.

\item Contrasting of the relations used in SRC tasks with regard to relations present in the  foundational ontology DOLCE, in the context of a domain-specific corpora.

\item Annotation of terms within sentences from a financial corpus with semantic relations, including composite semantic relations, and creation of a domain-specific test collection for relation classification.

\end{itemize}

The paper is organised as follows: Section \ref{sec:semrelclf} lists related work regarding the semantic relation annotation task.
 Section \ref{sec:critique} presents an analysis of current sets of semantic relations, and describes the relations provided by the foundational ontology DOLCE.
 Section \ref{sec:corpus-ana} describes the corpus-based analysis, followed by the conclusions and future work in Section \ref{sec:conclu}.

\section{The Semantic Relation Classification Task}\label{sec:semrelclf}

Semantic Relation Classification is usually framed under the context of a supervised classification problem.
 Best practices for creating relation inventories have been subject to much discussion \cite{seaghdha2007designing}.
 Inventories can either be organised under a hierarchical \cite{rosario2001classifying}, \cite{nastase2003exploring},
\cite{masolo2003wonderweb} or under a flattened approach \cite{moldovan2004models}.

The number of relations in a given inventory varies widely, ranging from binary classification \cite{lapata2002disambiguation} to 35 classes \cite{moldovan2004models} to open (inference-based) approaches \cite{sabou2008scarlet}.

There are several test collections for Semantic Relation Classification. Task 8 in SemEval 2010 \cite{hendrickx2009semeval} focuses on multi-way semantic relation classification between pairs of nouns.
 Nine relations with broad coverage were selected\footnote{Cause-Effect (CE), Instrument-Agency (IA), Product-Producer (PP), Content-Container (CC), Entity-Origin (EO), Entity-Destination (ED), Component-Whole (CW), Member-Collection (MC), Message-Topic (MT)}, with a focus on practical interest.
 Patterns were used to collect relation candidates from the web, which were then classified by two annotators.
 In the context of Distributional Semantics, BLESS \cite{baroni2011we} is a test collection which is designed to evaluate Distributional Semantic Models (DSMs) on the task of Semantic Relation Classification. BLESS provides a benchmark for evaluating the lexical semantic capabilities of DSMs: it provides \emph{concept, relation, relatum} triples for a large range of common concepts.
 There are five lexical semantic relations (\emph{co-hyponym}, \emph{hypernym}, \emph{meronym}, \emph{attribute} and \emph{event}) and three random relations (\emph{random-noun}, \emph{random-verb}, \emph{random-adjective}), which provide additional value for discriminativeness assessments.
 Some work has been done on SRC for specific domains, with a focus on the medical domain.
 Stephens et al. \shortcite{stephens2001detecting} distinguish 17 relations holding between genes.
 Rosario and Hearst \shortcite{rosario2001classifying} classify relations between noun compounds in the medical domain, while Rosario et al. \shortcite{rosario2002descent} undertake a similar endeavour using the MeSH hierarchy.
 Rosario and Hearst \shortcite{rosario2004classifying} explore SRC for biomedical texts, focusing on relations between treatments and diseases such as ``prevents'', ``cures'' or less specific effects.

\section{A Critique of Existing Sets of Semantic Relations}\label{sec:critique}

\subsection{SemEval-2010 Task 8} \label{sec:semeval-rel}

Although the Semeval-2010 Task 8 semantic relations set was developed with the aim of covering ``real word'' situations \cite{hendrickx2009semeval}, some of the constraints imposed to overcome the structural and lexical factors that can affect the truth of a relation, described next, can bring considerable limitations.
 In those cases,  it is necessary to identify other classes of semantic relations between terms covering other lexical categories.

\subsubsection{Focus on Nominals} The first point to be noted refers to the entities involved in the classification: the task focuses on semantic relations between pairs of nominals, that is, the relation arguments are only noun-phrases where the head is a common-noun.

\subsubsection{Locality Constraint} The data used in the Semeval classification task also relies on a locality constraint, which means that only nominal expressions considered ``local'' to one another were chosen, excluding relations whose arguments occur in separate sentential clauses.
 Although in a few cases a long distance between the arguments can indeed indicate the absence of a proper relation, in our financial data we note many sentences where the concepts are not local to one another, and nevertheless it is possible to assign a relation to them.
 For example, consider the pair ``debt'' and ``creditworthiness'' in Example \ref{ex:dist1}, or the concepts ``credit union'' and ``caisse populaire'' in Example \ref{ex:dist2}.

\begin{exe}
\ex \label{ex:dist1} ``Your debt problem won't go away, but your creditworthiness will.
''
\ex \label{ex:dist2} ``In Quebec 70 per cent of the population belongs to a caisse populaire, while in Saskatchewan close to 60 per cent belongs to a credit union''.

\end{exe}

In both cases, the concepts are located in different clauses within the sentences, but it is possible to identify a relation between them which could be \emph{indirect reference} and \emph{sibling concept}, respectively.
 In this case, no Semeval relation fits, and custom relations are necessary to better express the relationship.

\subsubsection{Focus on Concrete Relations} Although not stated as a constraint, most Semeval relations seems to refer specifically to physical objects.
 For example, the relation \emph{Content-Container (CC)} is described as ``An object is physically stored in a delineated area of space''.
 In \emph{Instrument-Agency (IA)}, \emph{Product-Producer (PP)}, \emph{Entity-Origin (EO)} and \emph{Entity-Destination (ED)}, all the mentioned examples involve physical objects as instruments, a material product being produced or a concrete objects physically moving to/from a place.
 This focus on concrete relations poses challenges to the classification of semantic relations within certain domains, since concepts representing abstract entities or quantitative/qualitative roles, such as ``credit'', ``debit'', ``investment'', ``demand'', ``profit'', ``interest'', ``capital'' or ``price'', to mention a few, are very frequent.

\subsubsection{Conditionals} Finally, the exclusion of conditional clauses also imposes unnecessary generality constraints.
 The Semeval task considers, for example, that in Example \ref{ex:semeval-conditions} the presence of the "bleach solution" inside the "bottle" is a situation being described as holding in a counterfactual hypothetical world, so it is not possible to assign a relation that can be seen as true regardless of hypothesis confirmation.

\begin{exe}
\ex \label{ex:semeval-conditions} ``Suppose you were given a bottle that contains 400 grams of a 3.0\% bleach solution.''
\end{exe}

Conditional clauses are frequent in many domains, for example within the financial domain.
 This domain involves many variables and frequently a scenario is being described based on them and the possible values they can assume.
 Therefore, \emph{condition} indeed seems to be a suitable relation between certain concepts, as in Example \ref{ex:condition}, where the relation arguments are ``term'' and ``bought''.

\begin{exe}
\ex \label{ex:condition} ``TIPS can be held to maturity and have a minimum term of ownership of 45 days if bought through TreasuryDirect''
\end{exe}

In the light of these limitations, adopting a richer conceptual meta-model, such as the one provided by the DOLCE ontology \cite{masolo2003wonderweb}, allow us to cover a broader range of categories instead of focusing only on physical objects, and consequently bring us a wider variety of relations to link those categories.
 Since all relations have a well defined domain and range, we can also ensure that they are valid for a given pair of concepts.
 Our analysis of the dataset has also shown that a complementary set of custom relations is of substantial importance to express the correct relationship between domain-specific concepts or even between concepts that, although being very common, interact among them in very domain-specific situations.
 In Section \ref{sec:dolce-rel} below, we therefore describe the DOLCE ontology and its relations.

\subsection{DOLCE relations} \label{sec:dolce-rel}

DOLCE (Descriptive Ontology for Linguistic and Cognitive Engineering) is an upper level ontology developed as a module of the WonderWeb library of foundational ontologies \cite{masolo2003wonderweb}.
 It has a clear cognitive bias, that is, it aims at capturing the ontological categories underlying natural language and human common sense.

 
The most fundamental distinction in DOLCE is that between \emph{endurants} and \emph{perdurants}. DOLCE relations are organised in a hierarchical structure.There are two toplevel relations: \emph{immediate-relation}, defined as a relation that holds without mediating individuals, and \emph{mediated-relation}, a relation that (implicitly) composes other relations.
 Two additional branches, namely \emph{immediate-relation-i} and \emph{mediated-relation-i}, cover all the inverse relations (only 4 relations do not have an inverse, and 14 relations have themselves as inverse, i.e., they are symmetric).

The \emph{immediate-relation} branch has 23 sub-relations at its second level, many of them being also subdivided into further levels.
 Among them are worth highlighting: \emph{part}, the most general meronym relation; \emph{participant}, the immediate relation holding between endurants and perdurants and which, through the sub-relations of its sub-relation \emph{functional-participant}, can define the role played by the endurant in the perdurant, for instance: \emph{patient}, \emph{target}, \emph{theme}, \emph{performed-by}, \emph{instrument}, \emph{resource}, etc.
; and \emph{references}, a relation holding between non-physical objects and any other kind of entity (including non-physical objects themselves), which can be seen as a type of association where the non-physical object carries some kind of information that involves the referenced entity.

The \emph{mediated-relation} branch has 25 sub-relations at its second level, with again some of them subdivided into further sub-relations.
 Among them are worth noting: \emph{co-participates-with}, a relation between two endurants participating in the same perdurant; \emph{generic-location}, a relation defining the physical or abstract location of an entity; and \emph{temporal-relation}, a relation between perdurants which, through its sub-relations, describe how two occurrences are related with respect to their temporal locations: \emph{precedes}, \emph{temporally-coincides}, \emph{temporally-includes}, \emph{temporally-overlaps}, etc.

 The relations having more generic classes as domain and range, that is, classes at higher levels in the hierarchy, proved to be more useful for the semantic annotation task (cp.
 Section \ref{sec:corpus-ana} below).
 As most of the relations have an inverse, it is almost always possible to assign a suitable property regardless of the arguments order, without the need to indicate the direction of the relation.
 
 DOLCE relations show to be a suitable set for SRC tasks because, as an upper level ontology, DOLCE aims at covering entities in any domain of knowledge. Since any entity can be mapped to a DOLCE high level category, it is always possible to find a relation (or a subset of candidate relations) between two entities, which will be the relation(s)
between their upper level DOLCE categories. When the relation is defined specifically for a class, it determines in a meaningful way what kind of relationship this class can have 
with another one. On the other hand, when the relation is inherited from an ancestor class, the kind of relationship can become too general. To address this issue and avoid the use of semantically vague relations, a small set of custom relations was proposed to complement the DOLCE relations set (cp. Section \ref{sec:dirrel}). Notwithstanding, this complementary set was designed to be as domain-independent as possible, in order to fit not only the context where it was defined, but to be also useful in any SRC task.

\section{Corpus-based Analysis}\label{sec:corpus-ana}

The analysis methodology presented in this section consists in the annotation of semantic relations with the help of a corpus. The corpus focuses on financial discourse and was crawled considering two types of discourse: glossaries and encyclopedic articles.
 
In the sections below, we describe the construction of our financial corpus including word pair selection and annotation (Section \ref{sec:corpus-construction}) and the extensive manual classification analysis (Section \ref{sec:manual-classif-ana}).

\subsection{Corpus Construction} \label{sec:corpus-construction}

We created a financial corpus by crawling two distinct types of sources: a) definitions, comprising three sources:
the Bloomberg financial Glossary\footnote{\url{http://www.isotranslations.com/resources/Bloomberg\%20Financial\%20Glossary.pdf}} (8324 definitions; 212,421 tokens),
SGM Glossary\footnote{\url{http://www.sapient.com/content/dam/sapient/sapientglobalmarkets/pdf/thought-leadership/SGM_Glossary_2014_final.pdf}} (1007 definitions; 43,638 tokens)
and Investopedia Definitions\footnote{\url{http://www.investopedia.com/terms/a/}} (15476 definitions; 2,462,801 tokens), b) articles from two sources: Investopedia\footnote{\url{http://www.investopedia.com/articles/pf/}} (5890 articles; 5,129,793 tokens)
and Wikipedia (articles on Investment\footnote{\url{https://en.wikipedia.org/wiki/Wikipedia:WikiProject_Investment}} and Finance\footnote{\url{https://en.wikipedia.org/wiki/Wikipedia:WikiProject_Finance}}; 8306 articles; 6,714,129 tokens). Overall, our corpus contains 14,580,803 tokens.

After the creation of the financial corpus, we selected word pairs for relation classification according to the following methodology: Splitting the corpus into sentences, the first word of the pair was randomly selected amongst the tokens in the sentence, with the only constraint that it was listed in one of the three financial glossaries.
Then, the second word was manually selected.
 The sentence context was preserved for manual classification analysis (see Section \ref{sec:manual-classif-ana} below).

\subsection{Manual Classification Analysis} \label{sec:manual-classif-ana}
Our semantic relation classification comprised 300 pairs of words, each associated with a sentence context (see Section \ref{sec:corpus-construction} above).
 First, for each pair, a class from the foundational ontology DOLCE \cite{masolo2003wonderweb} was assigned to both concepts.
 These classes represent the primary, highest level category that the concept belongs to.
 This concept-ontology class alignment was performed with the aid of the WordNet-DOLCE alignment \cite{gangemi2003sweetening}.
 For each concept, the correct sense and its corresponding DOLCE class were manually identified and assigned to it.
 For simplicity, all adjectives and adverbs were assigned the class \emph{quality}.

After classifying both concepts, it is possible to search for the most suitable relation between them, which is a property from DOLCE having the classes assigned to the concepts as domain and range.
 For example, if one concept represents an \emph{agent}, and the other one an \emph{action}, the possible relations between them could be \emph{performs}, meaning that the agent performs the action, or \emph{prescribes}, signifying that the agent does not perform the action him/herself, but somehow causes it to happen and to be performed by other agent(s).
 Besides the domain and range information, the sentence context where the concepts appear also helps to identify the correct relation.
 This also means that the relation assigned represents the relationship between those concepts in a particular sentence; the same pair of words could have different meanings and/or show a different kind of relationship in other sentence.
 When no suitable relation could be found in DOLCE, a new relation or a composite relation was suggested.
 When suggesting a new relation, we tried to make it as general as possible, that is, not too tied to a specific context, so it could be later reused by other concept pairs. The manual classification was performed by an expert in conceptual modelling and later independently reviewed by a second expert. 

Following this methodology, three scenarios occurred: (1) there was a direct relationship between the two concepts, so either a DOLCE relation or a custom suggested relation could be directly assigned to them; (2) there were no direct relations, but the concepts were indirectly related through other concepts, then a composition of (DOLCE or suggested) relations was drawn, building a path made of intermediate concept pairs linking the concepts; (3) no relation between the two concepts could be found at all, because they were too far away from each other in the same clause, or because they were in different clauses in a sentence, or in different sentences in a paragraph.
 In the final classification, 72.
67\% (218 pairs) of the pairs were assigned a direct relation, 24.67\% (74 pairs) were linked through an indirect relation, and only 2.66\% (8 pairs) were not classified.
 The classification results are summarised in Table \ref{tab:classif-results}.

 \begin{figure}
 \centering
 \captionof{table}{Relation classification results}
 \label{tab:classif-results}
 \begin{minipage}{\textwidth} 
 \centering
 \begin{tabular}{|c|c|c|c|c|c|c|}
 \hline
\textbf{Relation type} & \multicolumn{2}{|c|}{\textbf{DOLCE relations}} & \multicolumn{2}{|c|}{\textbf{Custom relations}} & \multicolumn{2}{|c|}{\textbf{Total}} \\
\hline
 &
\# of pairs & \% & \# of pairs & \% & \# of pairs & \% \\
\hline
Direct & 77 & 35.32 & 141 & 64.68 & 218 & 72.67 \\
\hline
Composite\footnote{The numbers refer only to the first pair in each composite relation chain} & 36 & 48.65 & 38 & 51.35 & 74 & 24.67 \\
\hline
Unclassified & - & -  & - & -  & 8 & 2.66 \\
 \hline
 \end{tabular}
 \end{minipage}
 \end{figure}

\subsubsection{Direct Relations}\label{sec:dirrel}
 For most concept pairs it was possible to assign a direct semantic relation.
 Out of the 218 pairs where this scenario occurred, 35.32\% (77 pairs) were assigned a DOLCE property as a semantic relation, and for 64.68\% (141 pairs) of the pairs no DOLCE property fit, so a \emph{suggested} \emph{relation} was assigned to each of them.
The suggested relations are listed in Table \ref{tab:suggested-relations}, and the DOLCE properties are well documented in the ontology itself\footnote{\url{http://www.loa.istc.cnr.it/old/DOLCE.html}}.

\begin{center}
\scriptsize
 \captionof{table}{Descriptions and examples of suggested semantic relations}
 \label{tab:suggested-relations}
 \begin{longtable}{|p{2cm}|p{6cm}|p{6.5cm}|}
 \hline
\textbf{Relation} & \textbf{Description} & \textbf{Example}\\
\hline
Common ownership & Both concepts have the same owner & Not only has the territory taken on increasing \emph{debt} in the 21st century but it has less \emph{revenue} coming in to pay that debt.\\
\hline
Condition & The existence or occurrence of one concept is conditioned by the existence or occurrence of the other concept, or by a broader condition involving that concept & If that's you, having a solid \emph{credit} \emph{history} can help you get funding for a start-up or establish a home-equity \emph{line} \emph{of} \emph{credit} to get your project off the ground.\\
\hline
Co-occurring qualifier & Both qualifiers occur at the same time in the same entity & These models are based upon \emph{historical} \emph{market} data.\\
\hline
Coreference & Syntactic reference between concepts, where one of them (usually a relative pronoun) refers to the other one & [It] is one of two Federal Reserve Bank of Cleveland \emph{branch} offices (the \emph{other} is in Cincinnati).\\
\hline
Correlated variation & Both concepts represent measures, and the variation in one of them affects the variation in the other one & It also decreases the value of the \emph{currency} - potentially stimulating exports and decreasing imports - improving the \emph{balance} \emph{of} \emph{trade}.\\
\hline
Destination & One concept is the destination of the other one, which can be an (physical or abstract) object itself, or an event causing some object to move towards it & Through LIFFE CONNECT, LIFFE took its \emph{market} to its \emph{customers} wherever they were in the world.\\
\hline
Indirect ownership & One concept is a part or a kind of representation of an agent or organisation, who/which has the ownership of the other concept & When Birmingham Midshires became \emph{part} of the Halifax in April 1999 it had savings balances of \pounds5.9 billion and \emph{mortgage} assets of \pounds9.2 billion.\\
\hline
Indirect qualifier & One concept is a quality of something that has the other concept as a part or as a direct quality & The Crummey letter qualifies the transfer for the annual \emph{gift} tax \emph{exclusion} \ldots.\\
\hline
Indirect reference & One concept makes some kind of reference to the other one, having other events and/or objects as intermediates & Characteristics and \emph{risk} types of human capital differ for different \emph{individuals}.\\
\hline
Indirect result & One concept indirectly produces the other one, having other events and/or objects as intermediates & The \emph{acquisition} created the largest provider of brokerage and \emph{investment} services in Greece.\\
\hline
Indirect target & One concept indirectly affects the other one through one or more events, which can also involve other (physical or abstract) objects & The \emph{firm} employs shareholder activism to push for structural changes in \emph{target} \emph{companies}.\\
\hline
Instantiation & One concept is an instance of a class represented by the other one & The \emph{FICO} \emph{score} is the most commonly used of the \emph{credit} \emph{scores}.\\
\hline
Membership & One concepts is a member of a group or organisation represented by the other one & In 2004, Mary Mitchell, the \emph{president} at the time, retired after a 60 year career at the \emph{bank}, starting as a teller in 1944.\\
\hline
Opposition & One concept is an antonym of the other one & \ldots and beggar thy neighbour policies that serve ``\emph{national} constituencies at the expense of \emph{global} financial stability''.\\
\hline
Ownership & One concept has the ownership of the other one & The \emph{lessor} is the legal owner of the \emph{asset}.\\
\hline
Qualifier & One concept is a quality of the other one & It's got speculators searching for \emph{quick} \emph{gains} in hot housing markets.\\
\hline
Represented in & One concept has some kind of (physical or abstract) representation expressed in/by the other one & All details of that \emph{transaction} are stored in the one-time \emph{code}.\\
\hline
Sibling concept & Both concepts belong to same category and play similar roles in a given context & Operating activities include net income, \emph{accounts} \emph{receivable}, \emph{accounts} \emph{payable} and inventory.\\
\hline
Source & One concept is the source of the other one, which can be an (physical or abstract) object itself, or an event causing some object to move from it & As of May 2014, AirHelp had raised a \$400000 seed \emph{round} from \emph{business} angels.\\
\hline
Specialisation & One concept is a more specific subconcept of the other one & If a \emph{value} other than \emph{market} \emph{value} is appropriate \ldots.\\
\hline
Theme component & One concept is something that demands complementary information to make clear what it is about, and the other one is a piece of the whole information & The \emph{downside} to this is that one \emph{review} doesn't tell a customer very much about the product.\\
\hline
Used for & Both concepts represent (physical or abstract) objects, and one is used as an instrument to accomplish the other & Look for receipts for medical costs not covered by \emph{insurance} or reimbursed by any other health plan , property taxes, and job-related and investment-related \emph{expenses}.\\
\hline
Value component & One concept represents a measure (something to which a value can be assigned), and the other one is something that, along with other parameters, determines its final value & Valuation of \emph{life} \emph{annuities} may be performed by calculating the actuarial \emph{present} \emph{value} of the future life contingent payments.\\
\hline
Affects & The existence or occurrence of one concept has some kind of effect on the other concept & If the option they have written gets \emph{exercised} several things can happen: for both put and call \emph{writers} if an option expires unexercised or is bought to close it is treated as a short-term capital gain.\\
\hline
  \end{longtable}
  \end{center}

Among the DOLCE relations, the most frequent ones are \emph{patient} and its inverse \emph{patient-of}, as well as \emph{target} and its inverse \emph{target-of}, covering around 42\% (32 pairs) of the pairs in this scenario. These relations refer to the association between events and the (abstract or physical) objects they affect. The \emph{patient} relation means that the object has a relatively static role in the event. \emph{Target} is a specialisation of \emph{patient}, and can be seen as an object to which an event is more intentionally directed.

This can give us an idea about the most frequent kind of concept that the events in this domain take as objects. The most common classes occurring as \emph{patient} or \emph{target} of an event are \emph{legal-possession-entity}, such as ``money'', ``loan'', ``shares'', ``income'' or ``investment'', \emph{description}, like ``deal'', ``trend'' or ``agreement'', and \emph{situation}, such as ``merger'', ``integration'' or ``asset management'', being affected by events like ``pay'', ``buy'', ``invest'', ``complete'', ``manage'' and ``deliver'', for example.

The suggested relations provide an abstract structural framework to express unnamed (implicit) relations between concepts within the text, without the need to commit to a domain-specific ontological model. Among the suggested relations, the most recurrent ones are \emph{qualifier}, \emph{indirect} \emph{target} and \emph{ownership}, accounting for 47.5\% (67 pairs) of the pairs in this scenario. The high frequency of the \emph{qualifier} relation can give us a hint about what concepts commonly modifies/are modified by other concepts. Adjectives like ``solvent'', ``failed'' and ``eligible'' are usually associated with \emph{social-roles}, like ``company'' or ``bank'', while nouns denoting \emph{legal-possession}-entities frequently modifies other \emph{legal-possession-entities}, specialising them, as in the pairs ``mortgage'' and ``line [of credit]'', and ``capital'' and ``account''.

The \emph{indirect} \emph{target} relation reinforces the high frequency of the ``affecting-affected entity'' pairs observed in the DOLCE-based classification, but in this case having some kind of intermediate between them, and also accepting (abstract or physical) objects, and not only events, as affecting entity. In this case, an event serves as intermediate, for example: ``accountant'' has as indirect target ``funds'', mediated by the event ``examination'', that is, ``accountant'' directly performs the action ``examination'', which in turn has as direct target ``funds''. Similarly, ``liquidator'' has as indirect target ``company'' through the event ``liquidation'', ``recruiters'' indirectly targets at ``candidate'' through ``hire'', and so on.

Another frequent suggested relation worth noting is \emph{ownership}, which is very recurrent between \emph{social-roles}, such as ``company'' and ``bank'', or \emph{socially-constructed-persons}, like ``employers'', ``sellers'' or ``manager'' as the owner (both classes denote \emph{roles}, the first being played by a juridical entity, and the second by a physical person), and \emph{legal-possession-entities}, such as ``assets'', ``funds'', ``insurance'', ``money'' and ``account'' as the owned entity.

\subsubsection{Relation Composition}\label{sec:relcomp}
When no direct relation between the two concepts could be found, the other concepts standing between them were analysed, and, instead of a single relation, a chain of concept pairs, each of them with its suitable direct relation, linked the two concepts from the original pair. Note that this scenario is different from the ones where direct, suggested relations such as \emph{indirect} \emph{target}, \emph{indirect} \emph{ownership} or \emph{indirect} \emph{qualifier}, for instance, were applied. In those cases, even having other events or objects as intermediates, a close relationship could be identified between the concepts. A composition of relations was necessary only when the only cohesive association from one concept to another is achieved by a direct mention of relation chains. 

Considering the 74 concept pairs where only indirect relations applied, the average length of the relations chain is 2.66, that is, this is the average number of concept pairs necessary to link the concepts, where the first pair contains one of the concepts and the last one contains the other. For example, in Example \ref{ex:indir}, no direct relation between ``type'' and ``month'' can be inferred, but, analysing the intermediate concepts, the following chain can be drawn: ``type [\emph{references}] financing, financing [\emph{used-in}] payments, payments [\emph{happens-at}] month''.

\begin{exe}
\ex\label{ex:indir} ``With another type of developer financing you make regular payments each month''
\end{exe}


The most common classes in this scenario are \emph{event} and \emph{quality}, which means that, even in a short sentence, sometimes a concept is not affected by an event at all, having only a relatively weak relation to the one that does. For qualities, the most probable reason is the distance between the concepts, as qualities are more likely to appear close to the concepts they qualify, having no meaningful relation with concepts far away within the sentence.

Regarding the relations in the compositions, 53.3\% (111 auxiliary pairs) of them were classified using DOLCE relations, and 43.7\% (86 auxiliary pairs) using suggested relations. Again, \emph{patient} and \emph{target}, and their inverses \emph{patient-of} and \emph{target-of} are predominant, but here the relation \emph{performs} also stands out. As all of these relations have \emph{event} as domain or range, we can infer that, when no apparent relation exists between the concepts, possibly an event can help to explain why they co-occur. Among the suggested relations, \emph{qualifier} and \emph{ownership} were the most frequent semantic relations, again, due to the high occurrence of concepts belonging to the categories \emph{quality}, what leads to the \emph{qualifier} relation, and \emph{social-role} and \emph{socially-constructed-person}, which, along with the also frequent category \emph{legal-possession-entity}, in this sample showed to be very likely to appear as the ``owner-owned entity'' pair.

\subsection{Correlation between Semantic Relations and Semantic Relatedness} \label{sec:relations-relatedness}

In order to further investigate the properties of the three relation categories \emph{direct}, \emph{composite}, \emph{unassigned} we correlate them in terms of their semantic relatedness scores. Two human annotators scored each of the 300 concept pairs for semantic relatedness on a scale from 0 (unrelated) to 10 (identical or highly related), where the average of their scores was taken as the final score of the concept pair. Note that the relatedness scoring, unlike the semantic relation assignment, was done without reference to the sentence context in order to obtain a general semantic relatedness assessment (replicating the methodology of \cite{finkelstein2001placing}).



If we consider the types of direct relations with regard to semantic relatedness, we find that the most highly related ones are \emph{Specialisation} (9.5; custom), \emph{Component-of} (9; DOLCE), \emph{Descriptive-place-of} (9; DOLCE), \emph{Product} (9; DOLCE), \emph{Use-of} (\emph{8}.\emph{5}; \emph{DOLCE}), \emph{Part-of} (8.25; DOLCE), \emph{Unit-of} (DOLCE; 8.25). In more general lexical semantic terms, they are instances of hyponymy (\emph{Specialisation}), meronymy (\emph{Component-of}, \emph{Part-of}), (abstract) location (\emph{descriptive-place-of}), and association (\emph{Unit-of}, \emph{Use-of}) and thus scored as highly related in our annotation schema.
The relations whose concepts on average display lowest relatedness are \emph{Happens-at} (3; DOLCE), \emph{Involves} (3.5; DOLCE), \emph{Result} (3.5; DOLCE), \emph{Source} (3.66; custom).\emph{Happens-at} has temporal characteristics, which do not necessitate high relatedness. \emph{Involves}, \emph{Result} and \emph{Source} have a low number of concept pairs in our data (one instance each for \emph{Involves} and \emph{Result}, three for \emph{Source}), which is why these results do not generalise.

\section{Conclusions and Future Work}\label{sec:conclu}

The semantic relation classification (SRC) task is a fundamental step in the construction of lightweight semantic models for Natural Language Processing applications.
 Current SRC tasks focus on very general relations that deal well with common sense data, but whose expressivity proves to be limited when applied to domain-specific information.
 We presented an analysis of the semantic relations from SemEval-2010 (task 8), a widely used relations set in SRC tasks, evaluating its coverage and ontological soundness to assess its suitability to domain-specific data.

Given the drawbacks identified in our evaluation and guided by a corpus-based analysis, we proposed a set of semantic relations made up by the properties of the foundational ontology DOLCE, complemented by a set of custom relations, and used it to classify a set of 300 pairs of terms from a financial dataset.
 As a result, besides the direct ontology-based relations, we introduced the concept of composite relations, a combination of one or more relations intended to link terms for which no direct relationship exists.
 The direct relations show us how the concepts interact and the composite relations help us to explain how terms that seem to be unrelated interact within a given context.

In addition to the manual relation classification, the pairs also received a score to indicate their semantic relatedness, independent of the context where they appear.
 Comparing the results of both classifications, we noted that pairs in a direct relationship have, on average, the highest semantic relatedness scores.
 The most predominant scenarios express how concrete or abstract objects are targeted by an event, are owned by an agent, or are modified/qualified by other objects.
 In contrast, pairs involved in a composite relationship present, on average, the lowest semantic similarity scores, showing that their relatedness is highly dependent on the context and can only be determined through a set of intermediate terms.

This initial classification shows that a conceptually well-grounded set of relations based on an ontological model can bring more expressivity and more flexibility for domain-specific data than that provided by the Semeval relations set.
 As future work, we intend to expand our analysis also to the correlation between contextual semantic and syntactic relations, as well as to extend our dataset, annotating a larger number of concept pairs and using this data to train an automatic classifier, capable of identifying semantic relations in large-scale corpora.

\section*{Acknowledgements}
\begin{minipage}[t]{.7\textwidth}
This work is in part funded by the SSIX Horizon 2020 project (grant agreement No 645425). Vivian S. Silva is a CNPq Fellow -- Brazil.
\end{minipage}%
\begin{minipage}[t]{.3\textwidth}
\raggedleft \raisebox{-.5\height}{\includegraphics[width=0.73in, height=0.4in]{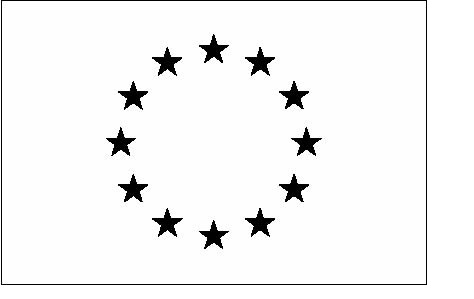}}
\end{minipage}

\bibliographystyle{acl}
\bibliography{coling2016}

\begin{thebibliography}{}

\bibitem[\protect\citename{Baroni and Lenci}2011]{baroni2011we}
Marco Baroni and Alessandro Lenci.
\newblock 2011.
\newblock How we blessed distributional semantic evaluation.
\newblock In {\em Proceedings of the GEMS 2011 Workshop on GEometrical Models
  of Natural Language Semantics}, pages 1--10. Association for Computational
  Linguistics.

\bibitem[\protect\citename{Finkelstein \bgroup et al.\egroup
  }2001]{finkelstein2001placing}
Lev Finkelstein, Evgeniy Gabrilovich, Yossi Matias, Ehud Rivlin, Zach Solan,
  Gadi Wolfman, and Eytan Ruppin.
\newblock 2001.
\newblock Placing search in context: The concept revisited.
\newblock In {\em Proceedings of the 10th international conference on World
  Wide Web}, pages 406--414. ACM.

\bibitem[\protect\citename{Gangemi \bgroup et al.\egroup
  }2003]{gangemi2003sweetening}
Aldo Gangemi, Nicola Guarino, Claudio Masolo, and Alessandro Oltramari.
\newblock 2003.
\newblock Sweetening wordnet with dolce.
\newblock {\em AI magazine}, 24(3):13.

\bibitem[\protect\citename{Hendrickx \bgroup et al.\egroup
  }2009]{hendrickx2009semeval}
Iris Hendrickx, Su~Nam Kim, Zornitsa Kozareva, Preslav Nakov, Diarmuid
  {\'O}~S{\'e}aghdha, Sebastian Pad{\'o}, Marco Pennacchiotti, Lorenza Romano,
  and Stan Szpakowicz.
\newblock 2009.
\newblock Semeval-2010 task 8: Multi-way classification of semantic relations
  between pairs of nominals.
\newblock In {\em Proceedings of the Workshop on Semantic Evaluations: Recent
  Achievements and Future Directions}, pages 94--99. Association for
  Computational Linguistics.

\bibitem[\protect\citename{Lapata}2002]{lapata2002disambiguation}
Maria Lapata.
\newblock 2002.
\newblock The disambiguation of nominalizations.
\newblock {\em Computational Linguistics}, 28(3):357--388.

\bibitem[\protect\citename{Masolo \bgroup et al.\egroup
  }2003]{masolo2003wonderweb}
Claudio Masolo, Stefano Borgo, Aldo Gangemi, Nicola Guarino, and Alessandro
  Oltramari.
\newblock 2003.
\newblock {WonderWeb} deliverable {D18} ontology library (final).
\newblock Technical report, IST Project 2001-33052 WonderWeb: Ontology
  Infrastructure for the Semantic Web.

\bibitem[\protect\citename{Moldovan \bgroup et al.\egroup
  }2004]{moldovan2004models}
Dan Moldovan, Adriana Badulescu, Marta Tatu, Daniel Antohe, and Roxana Girju.
\newblock 2004.
\newblock Models for the semantic classification of noun phrases.
\newblock In {\em Proceedings of the HLT-NAACL Workshop on Computational
  Lexical Semantics}, pages 60--67.

\bibitem[\protect\citename{Nastase and Szpakowicz}2003]{nastase2003exploring}
Vivi Nastase and Stan Szpakowicz.
\newblock 2003.
\newblock Exploring noun-modifier semantic relations.
\newblock In {\em Fifth international workshop on computational semantics
  (IWCS-5)}, pages 285--301.

\bibitem[\protect\citename{O'Seaghdha}2007]{seaghdha2007designing}
Diarmuid O'Seaghdha.
\newblock 2007.
\newblock Designing and evaluating a semantic annotation scheme for compound
  nouns.
\newblock In {\em Proc Corpus Linguistics}.

\bibitem[\protect\citename{Paperno \bgroup et al.\egroup }2016]{paperno2016}
Denis Paperno, Germ{\'{a}}n Kruszewski, Angeliki Lazaridou, Quan~Ngoc Pham,
  Raffaella Bernardi, Sandro Pezzelle, Marco Baroni, Gemma Boleda, and Raquel
  Fern{\'{a}}ndez.
\newblock 2016.
\newblock The {LAMBADA} dataset: Word prediction requiring a broad discourse
  context.
\newblock In {\em Proceedings of the 54th Annual Meeting of the Association for
  Computational Linguistics, {ACL} 2016, August 7-12, 2016, Berlin, Germany,
  Volume 1: Long Papers}.

\bibitem[\protect\citename{Rosario and Hearst}2001]{rosario2001classifying}
Barbara Rosario and Marti Hearst.
\newblock 2001.
\newblock Classifying the semantic relations in noun compounds via a
  domain-specific lexical hierarchy.
\newblock In {\em Proceedings of the 2001 Conference on Empirical Methods in
  Natural Language Processing}, pages 82--90.

\bibitem[\protect\citename{Rosario and Hearst}2004]{rosario2004classifying}
Barbara Rosario and Marti~A Hearst.
\newblock 2004.
\newblock Classifying semantic relations in bioscience texts.
\newblock In {\em Proceedings of the 42nd annual meeting on association for
  computational linguistics}, page 430. Association for Computational
  Linguistics.

\bibitem[\protect\citename{Rosario \bgroup et al.\egroup
  }2002]{rosario2002descent}
Barbara Rosario, Marti~A Hearst, and Charles Fillmore.
\newblock 2002.
\newblock The descent of hierarchy, and selection in relational semantics.
\newblock In {\em Proceedings of the 40th Annual Meeting on Association for
  Computational Linguistics}, pages 247--254. Association for Computational
  Linguistics.

\bibitem[\protect\citename{Sabou \bgroup et al.\egroup }2008]{sabou2008scarlet}
Marta Sabou, Mathieu d’Aquin, and Enrico Motta.
\newblock 2008.
\newblock Scarlet: semantic relation discovery by harvesting online ontologies.
\newblock In {\em European Semantic Web Conference}, pages 854--858. Springer.

\bibitem[\protect\citename{Stephens \bgroup et al.\egroup
  }2001]{stephens2001detecting}
Matthew~J Stephens, Mathew~J Palakal, Snehasis Mukhopadhyay, Rajeev~R Raje,
  Javed Mostafa, et~al.
\newblock 2001.
\newblock Detecting gene relations from medline abstracts.
\newblock In {\em Pacific Symposium on Biocomputing}, volume~6, pages 483--496.

\end{thebibliography}



\end{document}